\documentclass[preprint,review,3p,times,a4paper]{elsarticle}
\usepackage[linesnumbered,ruled,vlined, ruled,vlined]{algorithm2e}
\usepackage{amsmath}
\usepackage{lineno}
\usepackage{pgfplots}
\pgfplotsset{compat=1.18}
\usepackage{subcaption}
\usepackage{tikz}
\usepackage{xcolor}
\usepgfplotslibrary{fillbetween}
\usepackage{float}
\usepackage{tikz}
\usetikzlibrary{spy}
\usepackage{multirow,booktabs}

\journal{arXiv}
\begin{document}
\begin{frontmatter}
\title{B-LSTM-MIONet: Bayesian LSTM-based Neural Operators for Learning the Response of Complex Dynamical Systems to Length-Variant Multiple Input Functions}
\author[label2]{Zhihao Kong}
\author[label3]{Amirhossein Mollaali}
\author[label4]{Christian Moya}
\author[label2]{Na Lu}
\author[label3,label4]{Guang Lin\corref{core}}
	\ead{guanglin@purdue.edu}
    \cortext[core]{corresponding author}

 \affiliation[label2]{organization={Purdue University},
             addressline={Lyles School of Civil Engineering},
             city={West Lafayette},
             postcode={47906},
             state={IN},
             country={USA}}
 \affiliation[label3]{organization={Purdue University},
			 addressline={School of Mechanical Engineering},
		 	 city={West Lafayette},
	 		 postcode={47906},
			 state={IN},
			 country={USA}}
 \affiliation[label4]{organization={Purdue University},
			 addressline={Department of Mathematics},
		 	 city={West Lafayette},
	 		 postcode={47906},
			 state={IN},
			 country={USA}}    
\begin{abstract}
Deep Operator Network (DeepONet) is a recently introduced neural network framework that is designed to learn the nonlinear operators such as the solution operator arising from ordinary differential equations (ODEs) describing complex systems. Multiple-input deep neural operators (MIONet) extended DeepONet to allow multiple input functions in different Banach spaces. Unlike traditional neural networks requiring an equispaced grid, MIONet provides flexibility for the grid spacing of the training dataset and does not enforce constraints on the location of the output. However, MIONet has known limitations, such as the inputs having to be offline, i.e., we need to know the whole input over the fixed time domain, and the trained model cannot be applied to a testing dataset that has varying sequence lengths. These limitations prohibit its application in many practical problems, e.g., the real-time prediction of complex dynamic systems. In this work, we redesign the MIONet and propose using Long Short Term Memory (LSTM) to learn the neural operator from time-dependent data. This framework relaxes the constraints on the discretization of data and leverages the power of LSTM to learn memories from real-time data with variable lengths and variable time domains. Then we presented the factors affecting the learning performance, such as the extrapolation ability of the proposed algorithm. In addition, we provide the proposed framework with effective uncertainty quantification capabilities. We achieved this using the novel Bayesian method that samples from the posterior distribution of MIONet parameters. Building upon this, we developed a variant of MIONet that seamlessly blends the LSTM architecture with Bayesian uncertainty quantification, aptly named ``B-LSTM-MIONet.'' This integration not only leverages the temporal processing strengths of LSTM but also enriches it with the robustness of Bayesian methods. The result is a more powerful and reliable model, capable of handling complex datasets with a high degree of precision and confidence. Numerical experiments on autonomous Lorentz systems and non-autonomous pendulum systems demonstrate this framework can effectively learn the dynamic response of the underlying governing equations. The last example of a solar power generation system shows its decent performance on real-world data.

\end{abstract}
\end{frontmatter}

\section{Introduction} \label{sec:intro}
Recent advancements in scientific machine learning (SciML) have provided much more efficient tools to approximate complex dynamic systems compared to traditional high-fidelity numerical schemes. Rapid surrogate models derived from observational data now substantially reduce the computational cost to solve practical problems like solid mechanics \cite{brodnik2023perspective}, structural health monitoring \cite{li2021attention,lin2022long,liu2021multiscale}, field problem solutions \cite{peng2022rapid}, fault diagnosis \cite{chen2021graph,chen2022fault}, medical imaging \cite{wang2019weakly,leong2022surrogate}, autonomous driving \cite{wang2021review}, and power grid simulation \cite{moya2023approximating}\\
A significant challenge in current neural network surrogate models lies in their generalization capability. Addressing this, the foundational work \cite{chen1995universal} introduced Operator Learning, a novel method aimed at learning the mapping between different function spaces. Building on this, \cite{lu_learning_2021} developed the Deep Operator Neural Network (DeepONet), capable of being trained with limited datasets while minimizing generalization errors. This influential research has been applied in various domains, including the prediction of linear instability waves in high-speed boundary layers \cite{di2021deeponet}, forecasting power grid's post-fault trajectories \cite{moya2023deeponet}, learning nonlinear operators in oscillatory function spaces for seismic wave responses \cite{liu2021multiscale}, and analyzing nanoscale heat transport \cite{lu2022multifidelity}. Additionally, several advancements of DeepONet have been proposed, such as Bayesian DeepONet \cite{lin2021accelerated,moya2023bayesian}, DeepONet with proper orthogonal decomposition \cite{lu2022comprehensive}, multiscale DeepONet \cite{liu2021multiscale}, a neural operator with coupled attention \cite{kissas2022learning}, physics-informed DeepONet \cite{wang2021learning,moya2023dae}, and the multiple-input deep neural operators (MIONet) \cite{jin2022mionet}.\\
Notwithstanding its effectiveness, DeepONet has several recognized limitations. Initially, it often necessitates discretizing the input function \( u \) at predetermined locations for each sample. However, relying on a fixed set of discretization points for the inputs can be problematic, as these locations may vary across samples. Moreover, the standard DeepONet model does not incorporate temporal causality; that is, the current state of a system is influenced solely by its present state and historical data, not by future developments \cite{liu2022causality}. This absence of causality can lead to inaccuracies and physical inconsistencies in the model, failing to learn the inherent physics of the system.\\
These limitations have a crucial impact on the real-time prediction of complex dynamic systems, where the length of observational data increases with time. A few recent studies were made to address these limitations, for example, \cite{zhang2022belnet} introduced BelNet, a variant of DeepONet that is mesh-free, allowing arbitrary discretization positions for both training and testing dataset; \cite{lin2023learning} redesigned the DeepONet to learn the local solution operator, which maps the current system state and the input function in the near future to the next system state. While these methods provide flexibility on the sensor positions of input, they do not have an explicit mechanism to learn temporal physics. Meanwhile, some literature proposed to incorporate RNN with the DeepONet to learn temporal causalities from time-dependent input. \\
In their research, Liu et al. \cite{liu2022causality} developed the Causality-DeepONet, which employs zero-padding and a shifting convolution window to capture the causality in input data. However, this method encodes zeros through the branch network, subsequently merging them with the output from the trunk network. This process of introducing zero-padded data can inadvertently distort the final predictions, potentially leading to inaccurate or unintended results.\\
Park et al. \cite{park2023sequential} introduced the Sequential DeepONet (S-DeepONet), an innovative adaptation of the original DeepONet. This variant incorporates an autoregressive RNN branch for input encoding and decoding. The effectiveness of S-DeepONet was demonstrated through examples involving transient heat transfer and finite element deformation. This model notably achieved a reduction of average error by half across all testing samples compared to the standard DeepONet. However, the research focused on fixed time windows and initial values, leaving the model's response to dynamic systems with variable time lengths and initial conditions unaddressed.\\
Michalowska et al. \cite{michalowska2023neural} proposed a two-step method for considering temporal dependencies. In this approach, inputs are initially processed by a conventional DeepONet, followed by a subsequent RNN network for post-processing. This technique, utilizing a moving window, aims to overcome the limitations associated with fixed sensor positions, leading to what they refer to as ``short-interval one-shot predictions.'' While this method allows for recursive prediction generation, the reliance on a fixed time window may limit its capability to learn and represent long-term physical behaviors.\\
On the other hand, the performance of DeepONet can be limited to small and noisy datasets. To mitigate this, we introduced a framework known as Bayesian DeepONet (B-DeepONet) in \cite{lin2023b}, designed to yield accurate results even with noisy data. The training of B-DeepONet leverages the stochastic gradient replica exchange Langevin diffusion method, a technique we explored and refined in \cite{lin2023b,deng2020non,sahin2024deep}, as an alternative to traditional stochastic gradient-based training. B-DeepONet has demonstrated its utility in various applications, consistently delivering reliable outcomes. Furthermore, B-DeepONet not only offers dependable predictions through confidence intervals but also enhances efficiency by enabling faster training processes \cite{lin2023b}.\\
In this paper, we modify the original MIONet framework to learn the next-state response of a dynamic system based on its input, history of state, and current state. Inspired by recent advancements in Natural Language Processing (NLP) studies, the proposed framework learns from the memory of historical data instead of discretized input trajectory, so it can process real-time data with variable length. To train this framework, we first curate a training dataset by applying masks with varying lengths to the input function, then we employ the LSTM network to encode the input function into a memory space. The memory and the current state of the system are two branch nets, and the step size is the trunk net of the MIONet. The parameters of branch nets and trunk net are fused using the dot product to get the next system state. Such a data-driven operator framework can help forecast the near-future behavior of dynamic systems based on real-time data.\\
The objectives of this paper are summarized below.
\begin{enumerate}
  \item Approximate the solution operator of a complex system with variable input lengths: This effort is to create a deep learning framework that combines the MIONet and LSTM and learns the mapping between (i) the current state and temporal memory and (ii) the next state of the dynamic system. The inputs to such a framework are not required to have a constant length.
  \item Predict the dynamic response in unseen time domains: This effort is to extrapolate the solution operator to a different time domain than that of the training dataset, which enables medium or long-term projection for the dynamic response of the control system.
  \item Evaluate the robustness of the proposed neural operator: This effort is to evaluate the performance of the LSTM-MIONet for noisy datasets. Bayesian and stochastic gradient replica-exchange MCMC tools are used to quantify the uncertainty of test results.
\end{enumerate}
The contributions to achieving the objectives of this work are summarized below.
\begin{enumerate}
  \item In Section~\ref{sec:methods}, we present a reformulation of the complex dynamic system within a sequence-to-point context and propose an LSTM-enhanced MIONet framework that handles inputs with variable lengths.
\item We demonstrate the effectiveness of Bayesian and replica exchange Stochastic Gradient Langevin Diffusion (reSGLD) in handling the noise-polluted data and estimate the uncertainty of the test results in Section~\ref{sec:experiments}.
\item In Section~\ref{sec:experiments}, we also evaluate the proposed frameworks using cutting-edge models, including the Lorentz system, the pendulum system, and a power engineering application. Across all these experimental setups, the proficiency of the methods is consistently evident.
  \item We evaluate both the interpolation and extrapolation performance of the proposed framework in Section~\ref{sec:temporal}. The results demonstrate its versatility in the short and long term, and even for unseen (extrapolated) time domains.
\end{enumerate}
The remainder of this paper is structured as follows. Section~\ref{sec:statement} delineates the problem we aim to address and outlines our specific objectives. Following this, Section~\ref{sec:methods} provides an insightful review of existing neural operators, specifically focusing on DeepONet and MIONet, and introduces our proposed enhancement: the LSTM-enhanced MIONet algorithm. This section also explores the integration of Bayesian methods and the replica exchange Stochastic Gradient Langevin Diffusion (reSGLD) technique for Uncertainty Quantification (UQ). Subsequently, in Section~\ref{sec:experiments}, we test our algorithm through a series of numerical experiments to validate its efficacy. Section~\ref{sec:temporal} is devoted to examining how temporal parameters, such as the step size and extrapolation length, influence the performance of our algorithm. We summarize our findings of this paper in Section~\ref{sec:conclusion} and offer insights into potential avenues for future research.

\section{Problem Statement} \label{sec:statement}
Consider the complex dynamic system, whose state variable $x$ evolves according to an unknown system of governing equations. For example, $x$ may follow a system of non-autonomous ordinary differential equations (ODEs),
\begin{align}
    \frac{d}{dt}x(t)&=f(x(t),u(t)), t \in [t_0,t_0+T] 
\label{eq:govern}
\end{align}
where $x(t) \in \mathcal{X}_{[t_0:t]} \subseteq \mathbb{R}^{j}$ is the vector-valued state function, $u(t) \in \mathcal{U}_{[t_0:t]} \subseteq \mathbb{R}^{k}$ is an input function, $f \colon \mathcal{X}_{[t_0:t]} \times \mathcal{U}_{[t_0:t]} \mapsto \mathcal{X}_{[t_0:t]}$ is the unknown vector field, and $T \in \mathbb{R}$ is the maximum length of all instances of $x$. The Banach spaces $\mathcal{X}_{[t_0:t]}$ and $\mathcal{U}_{[t_0:t]}$ are dynamic in the time domain, i.e., their sizes expand when more data are available.\\
The non-autonomous dynamics describe the behavior of a complex dynamic system (\ref{eq:govern}). It enables input functions to have different lengths within the maximum time span $T$. Our goal is to create an accurate predictive model for the state $x$ such that their near-future dynamics can be studied.
\subsection{Real-Time Data}
\label{sec:realtime}
In some practical problems, such as the real-time system, on the one hand, we only have access to the historical data of a sequence and cannot see the future data in the inference stage, on the other hand, the data recorded for training is usually up to their maximum historical length. To handle such conflicting data formats, we design the \textit{mask} that recovers real-time data from recorded full-length historical data. Specifically, we denote $\mathcal{T}_{(t_0,t]}$ as the \textit{mask} for the input function $u$, and 
\begin{align*}
    \mathcal{T}_{(t_0,t]}(s)=
    \begin{cases} 
    1 & s \in  (t_0,t] \\
    0 & \text{otherwise}
    \end{cases}
\end{align*}
Then we apply the mask $\mathcal{T}_{(t_0,t]}$ to $u$ and get the expression of real-time input function:
\begin{align*}
    u_{(t_0,t]}:=u \cdot \mathcal{T}_{(t_0,t]}
\end{align*}
We note that $t$ is not necessarily equispaced or fixed at certain positions. The only requirement on $t$ is $t \in [t_0,t_0+T]$, i.e., the real-time sequential data $u_{(t_0,t]}$ can be derived from $u$ at arbitrary $t$ within the given time span.
\subsection{Modeling Objective}
This work aims to learn a \textit{local operator} that maps the current state, history of input function, and time spacing, to the next system state. Such \textit{local operator} is given by
\begin{align}
    \mathcal{F}(x(t),u_{(t_0,t]})(h(t)) = x(t+h(t)) \label{eq:2}
\end{align}
where $\mathcal{F}$ is the \textit{local operator}; $h(t) \in [0,h_{max}]$ is the time spacing between the current state $x(t)$ and the next state $x(t+h(t))$. The $h_{max}$ should intuitively be small, as the causality will decay over a long time horizon.
\section{Methods} \label{sec:methods}
DeepONet has shown efficacy across various applications. However, a notable constraint in existing neural operators, including DeepONet, is their design which is restricted to learning operators for functions within one Banach space, meaning the operator's input is confined to a single function. This restriction on the input space limits the scope for learning a broader array of valuable operators. For instance, it hinders the learning of operators like the ODE solution operator, which ideally maps from both the initial condition and the boundary condition to the ODE solution. This is because the initial condition and boundary condition are situated in two distinct domains (the initial domain and the boundary domain, respectively), thereby exceeding the current input capacity of these neural operators.\\
To overcome this limitation, MIONet is designed to accommodate multiple infinite-dimensional Banach spaces as the branch nets. MIONet becomes DeepONet with one branch net and one trunk net, so all the techniques developed for DeepONet in can be directly used for MIONet. However, DeepONet and MIONet are offline frameworks and they cannot be trained on real-time data with variable lengths. In this section, we first briefly review DeepONet and MIONet. Then, we propose LSTM enhanced MIONet (LSTM-MIONet) that combines the advantages of LSTM and MIONet for learning real-time data of dynamic systems.
\subsection{DeepONet}
DeepONet represents a linear, trainable approximation of the operator $G$, which functions to map the input function space to the output function space. This approximation is parameterized through a neural network $G_{\theta}$, consisting of two distinct sub-networks: the trunk net and the branch net. The branch net's role is to encode the input function $u$, while the trunk net focuses on encoding the query location $t$. DeepONet, in its essence, is a high-level framework that does not specify particular architectures for its branch and trunk nets. As an illustrative example, the work in \cite{lu_learning_2021} employed the basic Feed Forward Neural Network (FNN) for both these sub-networks.\\ 
Denote $u$ as the input function and $y$ as the output function defined on the domain $D \subset \mathbb{R}^{d}$:
\begin{align*}
    u: D \subset \mathbb{R}
\end{align*} 
\begin{align*}
    y: D' \subset \mathbb{R}
\end{align*}
The operator mapping from the input functions to the output function is
\begin{align*}
    G: u \mapsto y
\end{align*}
We define \( u_m \) as the discrete approximation of \( u \), sampled at \( m \) interpolation points \((t_1,...,t_m)\), that is, \( u_m\colon={u(t_1),...,u(t_m)} \). The branch net transforms the discrete input \( u_m \) into a vector of trainable coefficients \( b \in \mathbb{R}^p \). Conversely, the trunk net processes the query location to yield a set of trainable coefficients \( \mathcal{\varphi}  \in \mathbb{R}^p \). The final output of DeepONet is then derived by merging these trainable coefficients \( b \) with the trainable basis functions \( \mathcal{\varphi} \) through the application of the dot product, augmented by the scalar bias \( b_0 \in \mathbb{R} \):
\begin{align}
    G_{\theta}(u)(t)\colon=\sum_{i=1}^p b_i \cdot \mathcal{\varphi}_i+b_0 \label{eq:1}
\end{align}
Clearly, this DeepONet prediction is a one-shot; it requires knowledge of the input at all interpolation points $(t_1,...,t_m)$. This requirement, however, is not always met in practical applications, especially for real-time problems where the length of available data is increasing with time. Furthermore, DeepONet cannot be applied on off-time data, i.e., the data collected outside the time range where the DeepONet was trained. \\
To alleviate this constraint, we take a different approach in this paper. First, we prepare the training dataset with variable sequence length by applying random masks. Then, we train the branch net using LSTM architecture, which encodes the sequential inputs in trainable coefficients. The outputs of the branch net are eventually fused with the outputs of the trunk net following equation (\ref{eq:1}). Further details are introduced below.
\begin{figure}[h]
    \centering
    \includegraphics[width=0.9\textwidth]{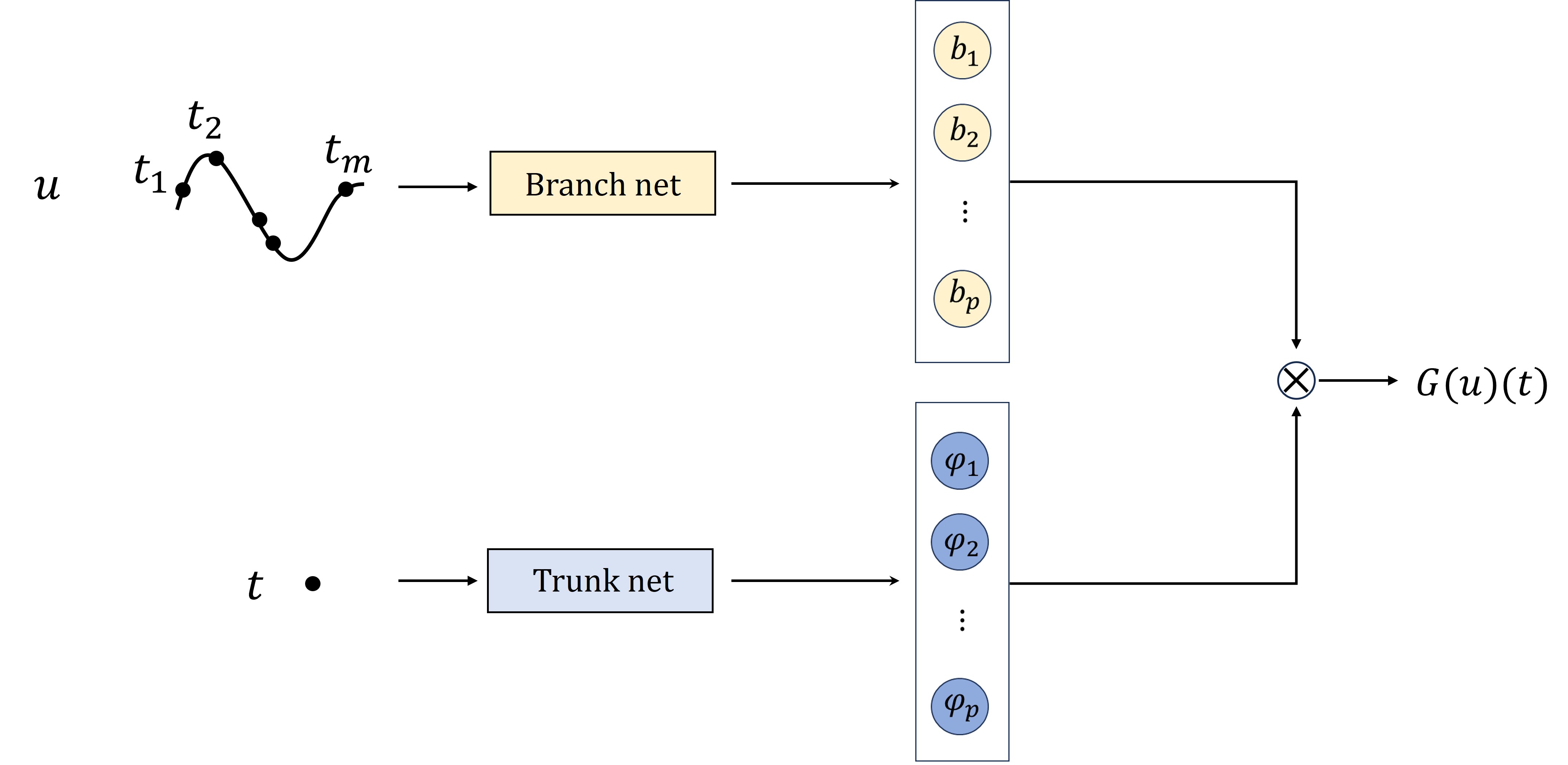}
    \caption{Architecture of DeepONet. This figure shows the DeepONet architecture, where the Branch net is used to encode the input function, and the Trunk net is used to encode the location. The integration of feature vectors from the Branch and Trunk nets is achieved through a dot product fusion, leading to the final output.}

    \label{fig:DON}
\end{figure}
\subsection{MIONet}
MIONet, introduced by \cite{jin2022mionet}, is designed for the learning of nonlinear operators that map across multiple Banach function spaces. Rooted in the universal approximation theorem by \cite{chen1995universal}, the conventional DeepONet is established for input functions within a single Banach space. MIONet expands the scope of DeepONet \cite{lu_learning_2021}, both in theory and practice, extending from a single Banach space to encompass multiple Banach spaces.\\
Denote $n$ input functions by $v_i$ for $i = {1, ..., n}$ with each defined on the domain $D_i \subset \mathbb{R}^{d_i}$:
\begin{align*}
    v_i: D_i \subset \mathbb{R}
\end{align*}
and the output function by $y$ defined on the domain $D' \subset \mathbb{R}^{d'}$:
\begin{align*}
    y: D' \subset \mathbb{R}
\end{align*}
Then, the operator mapping from the input functions to the output function is
\begin{align*}
    G: (v_1, ..., v_n) \mapsto y
\end{align*}
MIONet is designed to approximate the operator \( G \) through the utilization of \( n \) distinct branch nets alongside a singular trunk net. The \( i^{th} \) branch net is responsible for encoding the input function \( v_i \), while the trunk net is dedicated to encoding the coordinate input \( x \). The computation of MIONet's output is structured as follows:
\begin{align*}
    G_{\theta}(v_1, ..., v_n)(x) = \sum_{j=1}^p\left(\prod_{i=1}^nb_j^i(v_i) \cdot \mathcal{\varphi}_j\right)+b_0
\end{align*}
where $b_j^i$ represents the $j^{th}$ parameter of the $i^{th}$ branch net, $\mathcal{\varphi}_j$ is the $j^{th}$ parameter of the trunk net, and $b_0 \in \mathbb{R}$ is the scalar bias. 
\begin{figure}[h]
    \centering
    \includegraphics[width=0.9\textwidth]{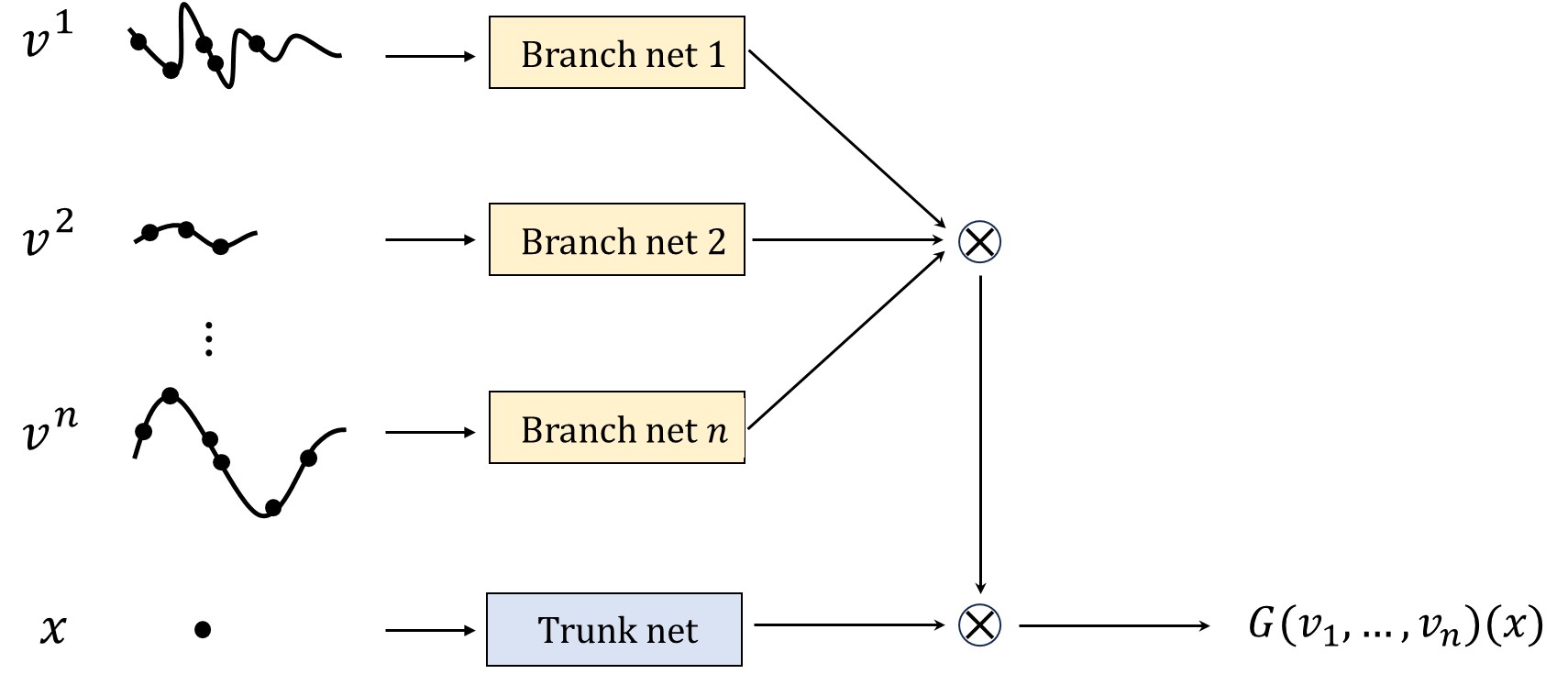}
    \caption{Architectural of MIONet. This diagram illustrates the MIONet structure, featuring multiple Branch nets that encode distinct input functions. The Trunk net is responsible for encoding spatial or location information. The integration of feature vectors from both the Branch nets and the Trunk net is achieved through a dot product operation.}
    \label{fig:MIONet}
\end{figure}
\subsection{The LSTM Enhanced MIONet (LSTM-MIONet)}
The vanilla DeepONet and MIONet use FNN to encode the sequential input data, and such input data remains identical for all interpolation positions. In consequence, the DeepONet and MIONet are not suitable for sequential data with variable lengths. To address this problem, we design the LSTM-MIONet $\mathcal{F}_\theta$, with the vector of trainable parameters $\theta$, to approximate the \textit{local operator} expressed in (\ref{eq:2}). This proposed LSTM-MIONet $\mathcal{F}_\theta$ framework consists of two branch nets and one trunk net, as illustrated in Figure~\ref{fig:1}.
\subsubsection{Non-autonomous Systems}
The first branch net employs FNN to map the current system state $x(t)$ to the trainable feature vector $b \in \mathbb{R}^p$. The second branch net encodes the history of input function $u_{(t_0,t]}$ to the trainable feature vector $\beta \in \mathbb{R}^p$ using an FNN-LSTM-FNN sandwich architecture. In the first layer, FNN maps the input vector instances at each time step of $u_{(t_0,t]}$ to the latent space; then in the second layer, LSTM encodes the latent sequences to a single hidden state or the memory; in the last layer, the memory of sequence is decoded into the output parameter space $\beta \in \mathbb{R}^p$ through another FNN. Meanwhile, the trunk net encodes the scalar step size $h(t)$ to the trainable feature vector $\varphi \in \mathbb{R}^p$ through FNN. The final output of $\mathcal{F}_\theta$ is computed by fusing all three trainable feature vectors using the following dot product:
\begin{align}
    \mathcal{F}_\theta(x(t),u_{(t_0,t]})(h(t))\colon=\sum_{i=1}^p b_i \cdot \beta_i \cdot \mathcal{\varphi}_i \label{eq:3}
\end{align}
\subsubsection{Autonomous Systems}
We recognize that for \textit{Autonomous Systems}, there does not exist an input function $u$, so we slightly modify the second branch net by substituting $u_{(t_0,t]}$ with $x_{(t_0,t]}$, i.e., the history of system state function $x$. The output trainable feature vector of $x_{(t_0,t]}$ is denoted by $\delta \in \mathbb{R}^p$. The modified LSTM-MIONet $\mathcal{F}^{\prime}_\theta$ for \textit{Autonomous Systems} is given by:
\begin{align}
    \mathcal{F}^{\prime}_\theta(x(t),x_{(t_0,t]})(h(t))\colon=\sum_{i=1}^p b_i \cdot \delta_i \cdot \mathcal{\varphi}_i \label{eq:4}
\end{align}
\begin{figure}[h]
    \centering
    \includegraphics[width=1\textwidth]{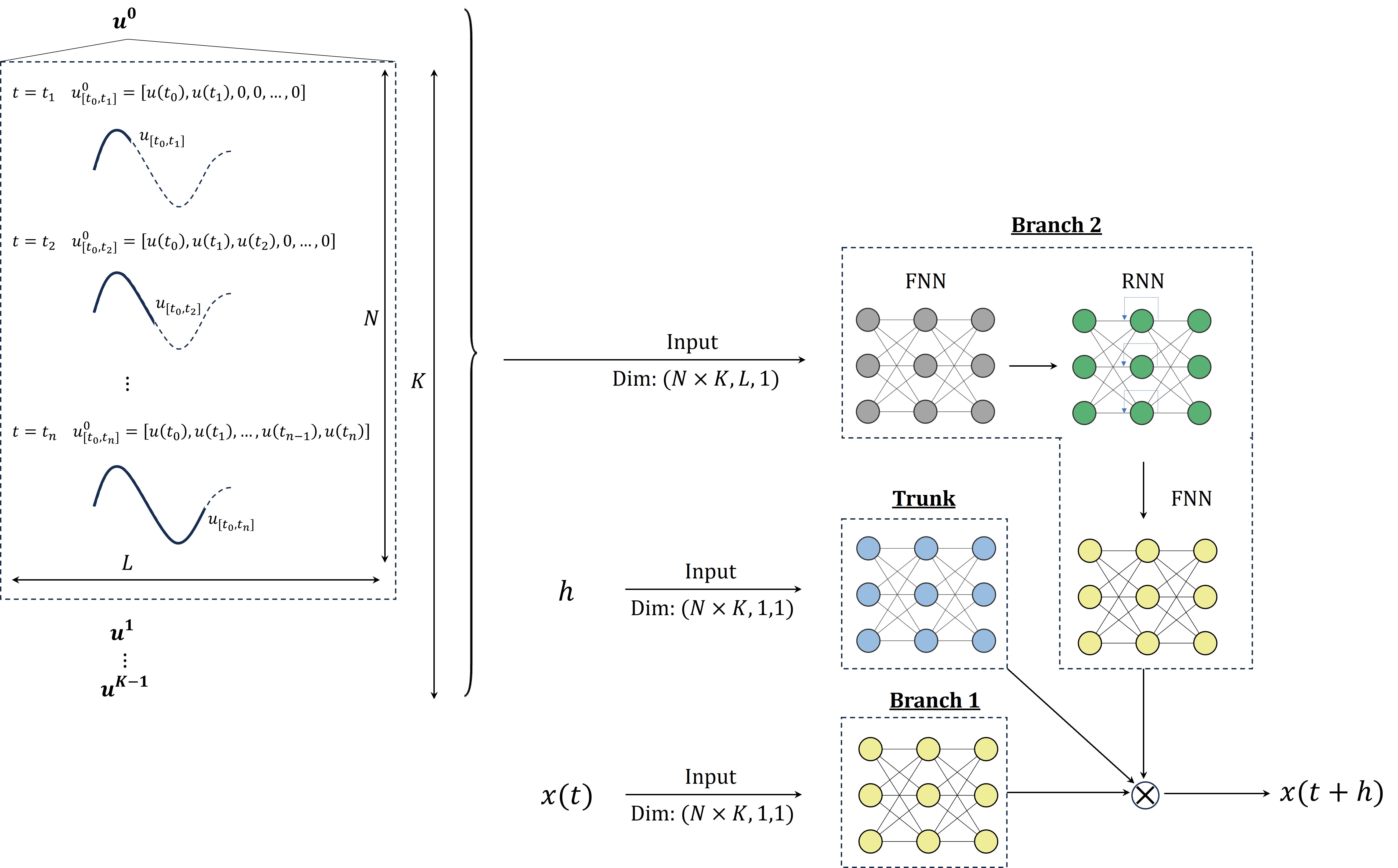}
    \caption{Architecture of LSTM-Enhanced MIONet. Branch net 1 encodes the system's current state, while Branch net 2 captures the system's temporal history. During training, input functions, ranging from \( t_0 \) to \( t_n \), are subjected to masks, adapting to different lengths. An LSTM network, combined with two FNNs, is employed to grasp the temporal dynamics of these inputs, accommodating their varied lengths. The Trunk net's role is to encode the interval between the current and forthcoming state of the complex system. The fusion of the current state, historical data, and temporal spacing through a dot product yields predictions for the system's subsequent state.}

    \label{fig:1}
\end{figure}

\subsection{Data Generation}
The LSTM-MIONet does not require the endpoint $t$ of a trajectory to be equispaced, while the trajectory itself has to be equispaced due to the nature of LSTM. Consider the discrete format of $t$ to be $(t_0,t_1,...,t_n)$, where $t_n \leq t_0+T$ and $\Delta \equiv t_{n}-t_{n-1}$, then the real-time input data are generated by applying masks to each sample of $u$ in the following form,
\begin{align*}
    u^i_{[t_0,t_1]}&=[u^i(t_0),u^i(t_1),0,...,0]\\
    u^i_{[t_0,t_2]}&=[u^i(t_0),u^i(t_1),u^i(t_2),0,...,0]\\
    \vdots\\
    u^i_{[t_0,t_n]}&=[u^i(t_0),u^i(t_1),...,u^i(t_n)]
\end{align*}
Each sequence here represents an evolution trajectory of $u^i$, and $n \geq 1$ is the non-zero length of each trajectory. As we stated in Section~\ref{sec:realtime}, the $t_n$ does not have to be continuous or equispaced, so the total number of real-time data generated from each $u^i$ is a random integer $1 \leq N \leq n$.\\
For \textit{Autonomous} system, we substitute the control function $u$ with state function $x$,
\begin{align*}
    x^i_{[t_0,t_1]}&=[x^i(t_0),x^i(t_1),0,...,0]\\
    x^i_{[t_0,t_2]}&=[x^i(t_0),x^i(t_1),x^i(t_2),0,...,0]\\
    \vdots\\
    x^i_{[t_0,t_n]}&=[x^i(t_0),x^i(t_1),...,x^i(t_n)]
\end{align*}
The training dataset is given by:
\begin{align*}
    \mathcal{D}_{train}=
    \begin{cases} 
    \{x^i(t_n),u^i_{(t_0,t_n]},h^i(t_n),x^i(t_n+h^i(t_n))\}_i^{N_{train}} & 
    \small{\text{\textit{for Non-Autonomous Systems}}} \\
    \{x^i(t_n),x^i_{(t_0,t_n]},h^i(t_n),x^i(t_n+h^i(t_n))\}_i^{N_{train}} & 
    \small{\text{\textit{for Autonomous Systems}}}
    \end{cases}
\end{align*}
\subsection{Bayesian Uncertainty Quantification}
While DeepONet and MIONet perform effectively in many applications, they would face challenges in scenarios where the training data is polluted with noise. This is a common situation in real-world applications and leads to poor generalization performance. The Bayesian DeepONet \cite{lin2023b,welling2011bayesian,chen2014stochastic}, in contrast, represents the prior for trainable parameters and utilizes replica exchange Stochastic Gradient Langevin Diffusion (reSGLD) to estimate the posterior distribution. This approach not only improves training convergence in noisy scenarios but also provides a more accurate mean predictive performance compared to vanilla DeepONets. Additionally, the Bayesian framework allows for the estimation of uncertainty in DeepONet predictions, enhancing the model's robustness and reliability in noisy environments. 
\subsubsection{Bayesian LSTM-MIONet}
In this work, we adapt the B-DeepONet proposed in \cite{lin2023b} to estimate the robustness of the proposed LSTM-MIONet to noisy data.\\
We consider the situation where the training targets for our operator are based on scattered and noisy observational data
\begin{align*}
    \Tilde{\mathcal{F}}(x(t),u_{(t_0,t]})(h(t)) = \mathcal{F}(x(t),u_{(t_0,t]})(h(t)) + \epsilon 
\end{align*}
where $\Tilde{\mathcal{F}}$ is the noisy measurements, the $\mathcal{F}$ is the true target values, $\epsilon$ is the independent Gaussian noise with zero and known standard deviation $\sigma$, i.e., $\sigma \sim \mathcal{N}(0,\sigma^2)$
Let $\mathcal{D}$ denote the noisy dataset of targets; we can then calculate the likelihood as
\begin{align*}
    P(\mathcal{D}|\theta)=\prod_{i=1}^N \frac{1}{\sqrt{2 \pi \sigma^2}}\exp\left(-\frac{(\mathcal{F}_i-\Tilde{\mathcal{F}}_i)^2}{2 \sigma^2}\right)
\end{align*}
To compute the posterior distribution, we use Bayes’ theorem, i.e.,
\begin{align*}
    P(\theta|\mathcal{D})=\frac{P(\mathcal{D}|\theta)P(\theta)}{P(D)} \propto P(\mathcal{D}|\theta)P(\theta)
\end{align*}
In practice, obtaining the posterior distribution via Bayes’ rule often presents computational and analytical challenges. Consequently, in our prior research \cite{lin2023b}, we resorted to approximating this posterior distribution by acquiring samples from it. To elaborate, we generated an M--ensemble of $\theta$ samples, represented as $\{\theta_k\}^M_{k=1}$, following the methodology described subsequently.
\subsubsection{M--ensemble Sampling}
Through introducing the replica exchange mechanism, we are able to efficiently sample from the posterior distribution, $P(\theta|\mathcal{D})$, thus enabling a more nuanced estimation of uncertainty in the predictions of the Bayesian MIONet. The replica exchange method involves running parallel chains of the Markov Chain Monte Carlo (MCMC) process to facilitate better exploration of the parameter space.\\
Specifically, two Langevin diffusions are employed within this framework. Each diffusion operates at a different ``temperature'' level, a mechanism that allows for an exploration of the parameter space at varying scales of granularity. One diffusion focuses on exploring the broader parameter space to escape local minima (high temperature), while the other is more focused on exploiting the local regions of the parameter space for finer details (low temperature). Periodically, these two diffusions engage in a swapping process, exchanging their states.\\
The incorporation of Gaussian noise in the Langevin diffusion process is a critical aspect of this approach. It introduces randomness into the trajectory of the MCMC, enabling the process to navigate through the parameter space more effectively. This randomness is controlled by the temperature parameter, with higher temperatures leading to larger jumps and thus a greater capacity for exploring the parameter space.\\
In the context of LSTM-MIONet, each model in the ensemble provides a set of predictions, allowing us to not only aggregate these predictions for a more robust estimate but also to assess the variability among the predictions.\\
Here we define the mean value and standard deviation of ensemble prediction results:
\begin{align*}
    \mu(x(t),u_{(t_0,t]})(h(t)) &= \frac{1}{M} \sum_{k=1}^{M} \mathcal{F}_{\theta_k}(x(t),u_{(t_0,t]})(h(t)) \\
    \sigma^2(x(t),u_{(t_0,t]})(h(t)) &= \frac{1}{M-1} \sum_{k=1}^{M} \left( \mathcal{F}_{\theta_k}(x(t),u_{(t_0,t]})(h(t)) - \mu(x(t),u_{(t_0,t]})(h(t)) \right)^2
\end{align*}
Then, we use the Prediction Interval Coverage Probability (PICP), denoted using $e$, as the metric to assess the effects of noisy parameters on our prediction intervals:
\begin{align*}
    e(x) = \frac{\text{\# of points of true value within 95\% confidence interval}}{\text{\# of points of true value}}
\end{align*}
This variability serves as a measure of uncertainty, offering insights into the confidence of the model's predictions under the influence of noisy training data. The ensemble approach, therefore, not only enhances the robustness of predictions but also provides a quantitative measure of uncertainty, which is crucial for decision-making in uncertain environments.
\section{Numerical Experiments} \label{sec:experiments}
To evaluate the LSTM-enhanced MIONet, we test it in three tasks: the autonomous Lorentz 63 system, the pendulum swing-up, and a power engineering application.
\subsection{Lorentz 63 System}
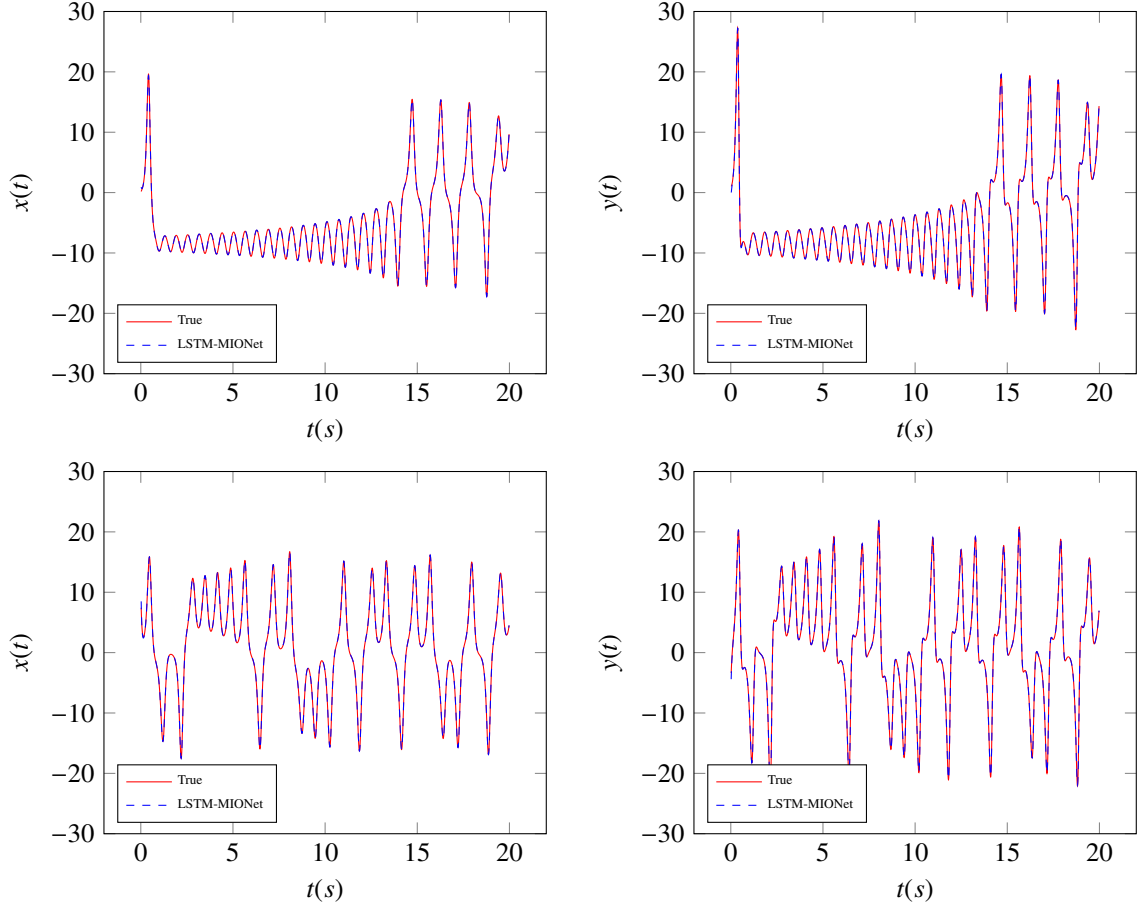
\begin{figure}[ht] 
    \begin{subfigure}[!bl]{0.45\textwidth}
        \begin{tikzpicture}[spy using outlines={magnification=2, width=1.5cm, height=4.5cm, connect spies}]
        \begin{axis}[width=\linewidth, xlabel={$t (s)$},ylabel={$x(t)$}, xmin=-2, xmax=22, ymin=-30, ymax=30, 
        xtick={0,5,...,20}, ytick={-30,-20,...,30}, legend pos=south west, legend style={font=\tiny}, legend cell align={left}]
        \addplot [smooth, red] table{data/lorenz_infer_x_single_true.txt}; 
        \addlegendentry{True}
        \addplot [dashed, blue] table{data/lorenz_infer_x_single_pred.txt}; 
        \addlegendentry{LSTM-MIONet}
        \end{axis}
        \end{tikzpicture}
    \end{subfigure}
    \hspace{5pt}
    \begin{subfigure}[!bl]{0.45\textwidth}
        \begin{tikzpicture}[spy using outlines={magnification=2, width=1.5cm, height=4.5cm, connect spies}]
        \begin{axis}[width=\linewidth, xlabel={$t (s)$},ylabel={$y(t)$}, xmin=-2, xmax=22, ymin=-30, ymax=30, 
        xtick={0,5,...,20}, ytick={-30,-20,...,30}, legend pos=south west, legend style={font=\tiny}, legend cell align={left}]
        \addplot [smooth, red] table{data/lorenz_infer_y_single_true.txt}; 
        \addlegendentry{True}
        \addplot [dashed, blue] table{data/lorenz_infer_y_single_pred.txt}; 
        \addlegendentry{LSTM-MIONet}
        \end{axis}
        \end{tikzpicture}
    \end{subfigure}\\
    \begin{subfigure}[!bl]{0.45\textwidth}
        \begin{tikzpicture}[spy using outlines={magnification=2, width=1.5cm, height=4.5cm, connect spies}]
        \begin{axis}[width=\linewidth, xlabel={$t (s)$},ylabel={$x(t)$}, xmin=-2, xmax=22, ymin=-30, ymax=30, 
        xtick={0,5,...,20}, ytick={-30,-20,...,30}, legend pos=south west, legend style={font=\tiny}, legend cell align={left}]
        \addplot [smooth, red] table{data/lorenz_infer_x_single_2_true.txt}; 
        \addlegendentry{True}
        \addplot [dashed, blue] table{data/lorenz_infer_x_single_2_pred.txt}; 
        \addlegendentry{LSTM-MIONet}
        \end{axis}
        \end{tikzpicture}
    \end{subfigure}
    \hspace{5pt}
    \begin{subfigure}[!bl]{0.45\textwidth}
        \begin{tikzpicture}[spy using outlines={magnification=2, width=1.5cm, height=4.5cm, connect spies}]
        \begin{axis}[width=\linewidth, xlabel={$t (s)$},ylabel={$y(t)$}, xmin=-2, xmax=22, ymin=-30, ymax=30, 
        xtick={0,5,...,20}, ytick={-30,-20,...,30}, legend pos=south west, legend style={font=\tiny}, legend cell align={left}]
        \addplot [smooth, red] table{data/lorenz_infer_y_single_2_true.txt}; 
        \addlegendentry{True}
        \addplot [dashed, blue] table{data/lorenz_infer_y_single_2_pred.txt}; 
        \addlegendentry{LSTM-MIONet}
        \end{axis}
        \end{tikzpicture}
    \end{subfigure}
\caption{Comparison of LSTM-MIONet prediction with the actual trajectory of the autonomous Lorentz system’s state $x = (x(t), y(t))^{\top}$ (left is $x(t)$ and right is $y(t)$) for the initial condition $(x(0), y(0), z(0)) = (0, 1, 1)$ (\textit{upper row}) and $(x(0), y(0), z(0)) = (10, -5, 20)$ (\textit{bottom row}) within the partition $\mathcal{P} \in [0,20]$ (s) of constant time spacing $h = 0.01$.}
\label{fig:2}
\end{figure}

\begin{figure}[ht]
    \centering
    \includegraphics[width=0.6\textwidth]{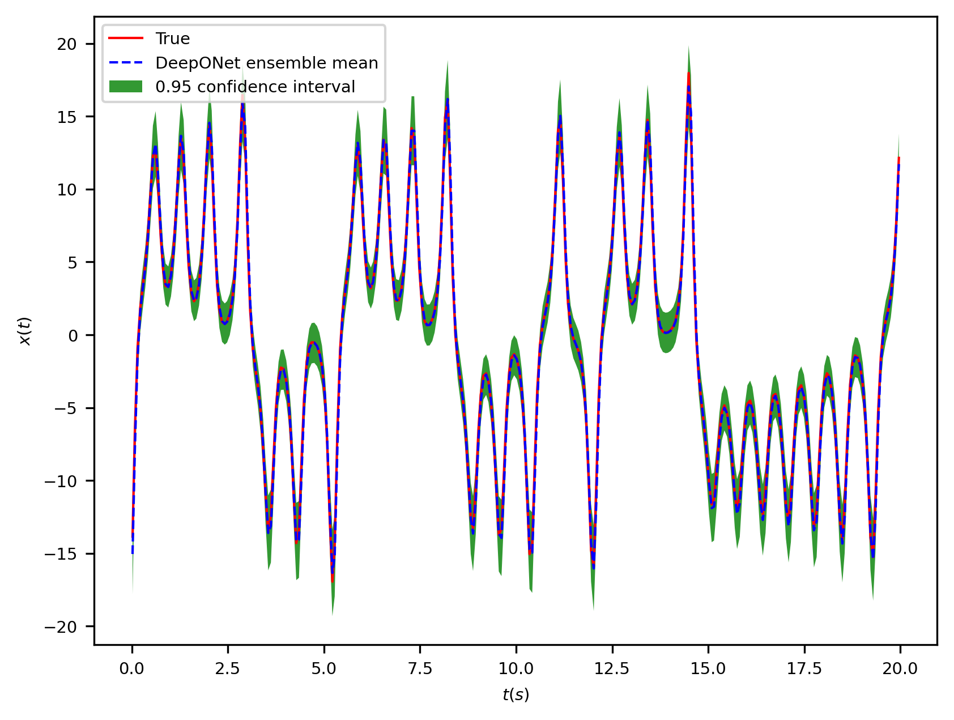}
    \caption{The confidence interval of B-LSTM-MIONet for Lorentz problem. The full trajectory of true data is within the confidence interval of ensemble prediction results.}
    \label{fig:UQ-lorentz}
\end{figure}

To evaluate the proposed framework, we first consider the autonomous and chaotic Lorentz 63 system with the following dynamics:
\begin{align*}
    \dot{x}&=\sigma(y-x) \\
    \dot{y}&=x(\rho-z)-y \\
    \dot{z}&=xy-\beta z 
\end{align*}
with parameters $\sigma=10$, $\rho=28$, and $\beta=8/3$. Notice that, compared to non-autonomous systems, autonomous systems do not have the input function $u$ as the input of the LSTM network, so we follow the modified networks defined in \ref{eq:4}.\\
The training dataset for this autonomous system is provided as a collection of quartets data $\{x(t_n),x_{(t_0,t_n]},h(t_n),x(t_n+h(t_n))\}$. We first sample 5000 initial values $x(0)$ from the state space $\mathcal{X}_0\colon=[-17,20] \times [-23,28] \times [0,50]$. Then, we solve the Lorentz ODE system on $[0,T]$ at the fixed step size $\Delta=0.01$ (s) using Runge-Kutta(RK-4), where $T=20$ (s). Let $x$ denote the solution functions. To produce $x_{(t_0,t_n]}$, we replicate $x$ for 4 times and apply the \emph{mask} $\mathcal{T}_{(t_0,t_n]}$ to each $x$, where $t_n$ is uniformly taken from $(0,T-h_{max})$. The local time spacing $h(t_n)$ is sampled from $h(t_n) \in [0,h_{max}]$, where $h_{max}=0.02$ (s). The next system state $x(t_n+h(t_n))$ is interpolated between adjacent $x(t_n)$ points. The masked solution function $x_{(t_0,t_n]}$, local step size $h(t_n)$, and current system state $x(t_n)$ are the inputs of the neural network. The next system state $x(t_n+h(t_n))$ is the output. In total, we have $N_{train}=20000$, i.e., 20000 sub-sequences of $x(t)$ with various initial values and lengths, corresponding to 20000 sets of current and next states.\\
The trained LSTM-MIONet is used to predict the response of the autonomous Lorentz system over the time-domain $[0,20]$ (s), with random initial conditions $x(0) \in \mathcal{X}_0$, with the fixed step size $\Delta=0.01$ (s) and the fixed time spacing $h=0.01$ (s).\\
Figure~\ref{fig:2} presents a comparative analysis of the LSTM-MIONet predictions against the actual behavior of the Lorentz system's state variables \( x(t) \) and \( y(t) \). The findings reveal a remarkable coincidence between the predictions of the proposed approach and the true data, underscoring its effectiveness even amidst the Lorentz system's inherent chaos.\\
The results of UQ using reSGLD are shown in Figure~\ref{fig:UQ-lorentz}. The number of ensembles is $M=300$. As shown in Table~\ref{tab:UQ-lorentz}, the $L_2$ relative error of the mean prediction versus the true trajectory is 5.83~\%. The PICP is 100~\%, i.e., the confidence interval covers all the true values. 
To test the predictive power of the proposed framework, we compute the average and standard deviation (st. dev.) of the $L_2$ - relative error between the predicted and actual response of the Lorentz system to 100 initial conditions sampled from $\mathcal{X}_0$. We replicate each $x(t)$ 200 times to generate 200 sub-sequences with varying lengths, so the $L_2$ relative error is calculated at 200 state points on each $x(t)$ trajectory. In total, we have the $L_2$ vector of size 100 corresponding to the 100 initial conditions. Results are reported in Table~\ref{tab:tb1}.\\
\begin{table}[ht]
    \centering
    \begin{tabular}{p{3cm}p{2cm}p{2cm}}
        \toprule
         & $x(t)$ & $y(t)$  \\
        \midrule
         mean $L_2$ & 1.29\% & 1.13\%\\
         st.dev. $L_2$ & 0.93\% & 0.70\%\\
         \bottomrule
    \end{tabular}
    \caption{The average and standard deviation (st.dev.) of the $L_2$ - relative error between the predicted and actual response trajectories of the Lorentz system within the partition $\mathcal{P} \in [0,20]$ (s) of constant time spacing $h=0.01$ to 100 initial conditions sampled from the set $\mathcal{X}_0\colon=[-17,20] \times [-23,28] \times [0,50]$.}
    \label{tab:tb1}
\end{table}
\begin{table}[ht]
    \centering
    \begin{tabular}{p{3cm}p{2cm}p{2cm}}
        \toprule
         & $x(t)$ & $y(t)$ \\
        \midrule
         $L_2$ & 5.83\% & 8.33\% \\
         PICP & 100.00\% & 97.00\% \\
         \bottomrule
    \end{tabular}
    \caption{The $L_2$ relative error and PICP of the example ensemble results of Lorentz system.}
    \label{tab:UQ-lorentz}
\end{table}
\subsection{Pendulum Swing-Up}
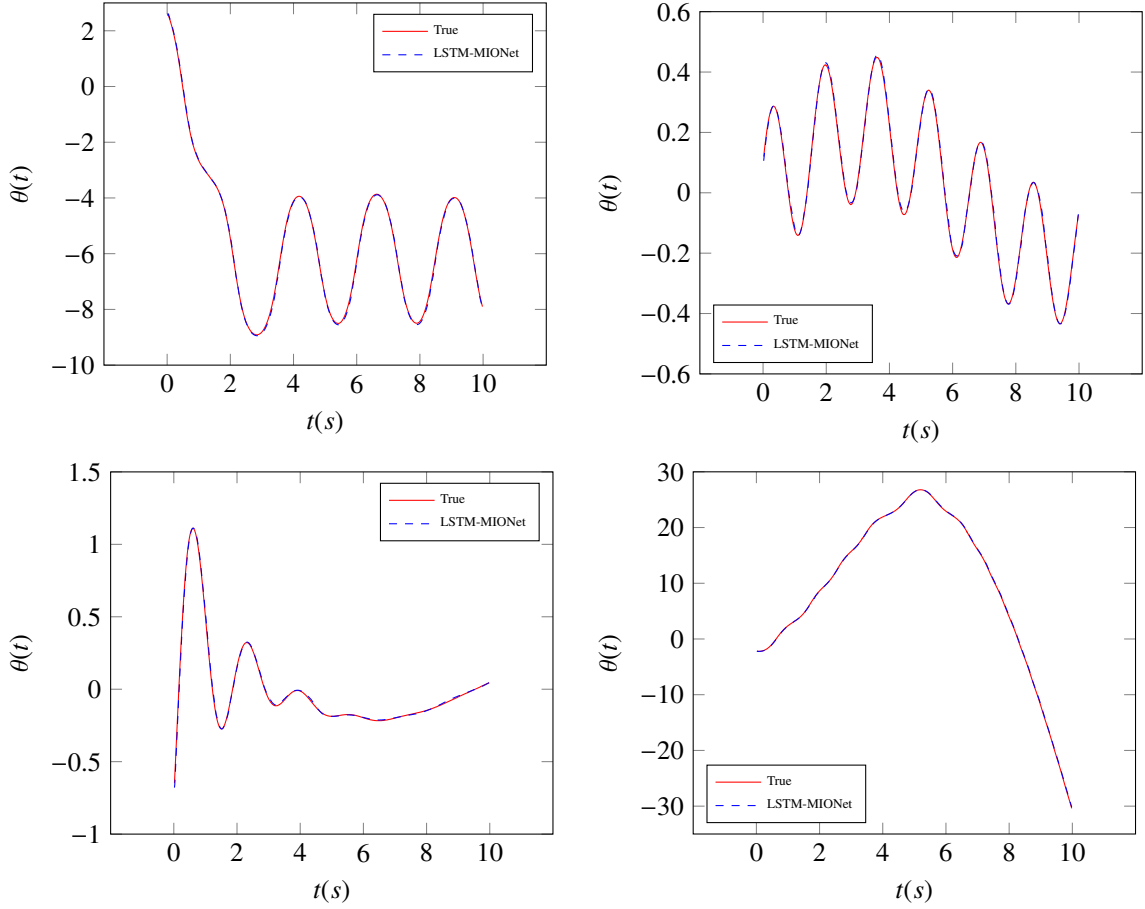
\begin{figure}[ht] 
    \begin{subfigure}[!bl]{0.45\textwidth}
        \begin{tikzpicture}[spy using outlines={magnification=2, width=1.5cm, height=4.5cm, connect spies}]
        \begin{axis}[width=\linewidth, xlabel={$t (s)$},ylabel={$\theta(t)$}, xmin=-2, xmax=12, ymin=-10, ymax=3, 
        xtick={0,2,...,10}, ytick={-10,-8,...,3}, legend pos=north east, legend style={font=\tiny}, legend cell align={left}]
        \addplot [smooth, red] table{data/pendulum_infer_theta_single_u_single_mix_true.txt}; 
        \addlegendentry{True}
        \addplot [dashed, blue] table{data/pendulum_infer_theta_single_u_single_mix_pred.txt}; 
        \addlegendentry{LSTM-MIONet}
        \end{axis}
        \end{tikzpicture}
    \label{subfig:mix}
    \end{subfigure}
    \hspace{5pt}
    \begin{subfigure}[!bl]{0.45\textwidth}
        \begin{tikzpicture}[spy using outlines={magnification=2, width=1.5cm, height=4.5cm, connect spies}]
        \begin{axis}[width=\linewidth, xlabel={$t (s)$},ylabel={$\theta(t)$}, xmin=-2, xmax=12, ymin=-0.6, ymax=0.6, 
        xtick={0,2,...,10}, ytick={-0.6,-0.4,...,0.6}, legend pos=south west, legend style={font=\tiny}, legend cell align={left}]
        \addplot [smooth, red] table{data/pendulum_infer_theta_single_chris_osc_true.txt}; 
        \addlegendentry{True}
        \addplot [dashed, blue] table{data/pendulum_infer_theta_single_chris_osc_pred.txt}; 
        \addlegendentry{LSTM-MIONet}
        \end{axis}
        \end{tikzpicture}
    \label{subfig:sin}
    \end{subfigure}\\
    \begin{subfigure}[!bl]{0.45\textwidth}
        \begin{tikzpicture}[spy using outlines={magnification=2, width=1.5cm, height=4.5cm, connect spies}]
        \begin{axis}[width=\linewidth, xlabel={$t (s)$},ylabel={$\theta(t)$}, xmin=-2, xmax=12, ymin=-1, ymax=1.5, 
        xtick={0,2,...,10}, ytick={-1,-0.5,...,1.5}, legend pos=north east, legend style={font=\tiny}, legend cell align={left}]
        \addplot [smooth, red] table{data/pendulum_infer_theta_single_cos_true.txt}; 
        \addlegendentry{True}
        \addplot [dashed, blue] table{data/pendulum_infer_theta_single_cos_pred.txt}; 
        \addlegendentry{LSTM-MIONet}
        \end{axis}
        \end{tikzpicture}
    \end{subfigure}
    \hspace{5pt}
    \begin{subfigure}[!bl]{0.45\textwidth}
        \begin{tikzpicture}[spy using outlines={magnification=2, width=1.5cm, height=4.5cm, connect spies}]
        \begin{axis}[width=\linewidth, xlabel={$t (s)$},ylabel={$\theta(t)$}, xmin=-2, xmax=12, ymin=-35, ymax=30, 
        xtick={0,2,...,10}, ytick={-30,-20,...,30}, legend pos=south west, legend style={font=\tiny}, legend cell align={left}]
        \addplot [smooth, red] table{data/pendulum_infer_theta_single_cos_2_true.txt}; 
        \addlegendentry{True}
        \addplot [dashed, blue] table{data/pendulum_infer_theta_single_cos_2_pred.txt}; 
        \addlegendentry{LSTM-MIONet}
        \end{axis}
        \end{tikzpicture}
    \end{subfigure}
\caption{Comparison of LSTM-MIONet prediction with the actual trajectory of the non-autonomous pendulum system’s state $\theta(t)$ for (i) $u \sim \mathcal{G}$ (\textit{upper left}) (ii) $u=\sin(t/2)$ (\textit{upper right}) (iii) $u=\cos(t/2) - \dot{\theta}/2$ (\textit{bottom left}) and (iv) $u=\cos(t/2) - t/5$ (\textit{bottom right}) within the partition $\mathcal{P} \in [0,10]$ (s) of constant time spacing $h = 0.01$ (s).}
\label{fig:3}
\end{figure}

\begin{figure}[ht]
    \centering
    \includegraphics[width=0.6\textwidth]{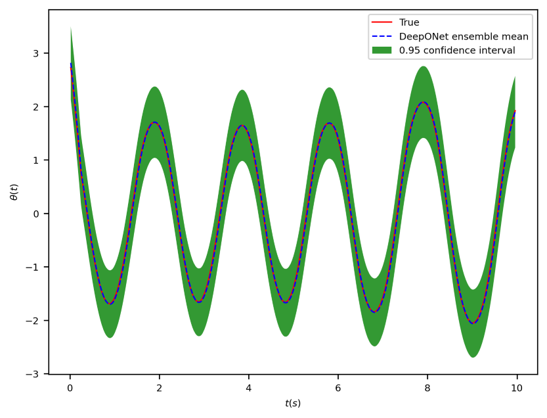}
    \caption{The confidence interval of B-LSTM-MIONet for pendulum problem. The full trajectory of true data is within the confidence interval of ensemble prediction results.}
    \label{fig:UQ-pendulum}
\end{figure}

Let us now consider the following pendulum swing-up control system:
\begin{align}
    \ddot{\theta}\left(\frac{1}{4}ml^2+I\right)+\frac{1}{2}mlg\sin{\theta}=u(t)-b\dot{\theta} \label{eq:5}
\end{align}\\
where \( x=(\theta,\dot{\theta})^{\top} \) represents the state vector in the space \(\mathcal{X}\), with \(\theta\) denoting the angle of the pendulum, \(\dot{\theta}\) its angular velocity, and \( u(t)\in \mathcal{U} \) the exerted control torque. Parameters of the pendulum include a mass of \( m=1 \) (kg), a length of \( l = 1 \) (m), a moment of inertia around its midpoint given by \( I =\frac{1}{12}ml^2 \), and a friction coefficient of \( b = 0.01 \) (sNm/rad).\\
For the \textit{Non-Autonomous} control task (pendulum systems), we generate the training dataset $\mathcal{D}_{train}$ following Figure~\ref{fig:1} and the procedures introduced here. We first uniformly take $K$ samples of the input function $u$ from the Gaussian random field $\mathcal{G}(0,k_l(x_1,x_2))$, where $k_l(x_1,x_2)$=exp$(-||x_1-x_2||^2/2l^2)$ is the radial-basis function covariance kernel \cite{lu_learning_2021} with the kernel length $l=0.01$. Then, we (i) discretize $u$ on the time positions $(t_0,t_1,...,t_n)$ (ii) replicate each $u$ for ${L}$ times, where ${L}$ represents the number of sub-sequences we will produce from each sample of $u$, and (iii) apply the mask $\mathcal{T}_{(t_0,t_n]}$ to $u$ and we have $u_{(t_0,t_n]}=u \cdot \mathcal{T}_{(t_0,t_n]}$, where $t_n$ is uniformly sampled from $(t_0,t_0+T]$. After sampling $u_{(t_0,t_n]}$ from the given function space, we solve the current system state $x(t_n)$ from the input $u_{(t_0,t_n]}$ and solve the next state $x(t_n+h(t_n))$, i.e., the reference solution state, from the $u_{(t_0,t_n+h(t_n)]}$ using  RK-4. The initial values $x(t_0)$ for RK-4 integration are uniformly sampled from the state space $\mathcal{X}_0$. In total, the sample size is given by $N_{train} = K \times L$.\\
To learn the dynamic response $x=(\theta,\dot{\theta})^{\top}$ of the pendulum system from the control function $u$, we train the LSTM-MIONet using $N_{train}=50000$ samples, each consisting of a quartet $\{x(t_n),u_{(t_0,t_n]},h(t_n),x(t_n+h(t_n))\}$.
We first sample 5000 initial values $x(0)$ from the state space $\mathcal{X}_0 \colon = \{ \theta,\dot{\theta} \colon \theta \in [-\pi,\pi], \dot{\theta} \in [-8,8] \}$. Meanwhile, we sample 5000 control functions $u$ from the Gaussian random field, i.e., $u \sim \mathcal{G}$. Then, we solve the pendulum system's response $x$ on partition $\mathcal{P} \in [0,T]$ (s) at the fixed step size $\Delta=0.01$ (s) using RK-4, where $T=10$ (s). To produce $u_{(t_0,t_n]}$, i.e., the sub-sequence of $u$ with variable length, we replicate $u$ for 10 times and apply the mask $\mathcal{T}_{(t_0,t_n]}$ to each $u$, where $t_n$ is uniformly taken from $(0,T-h_{max})$ (s). The local time spacing $h(t_n)$ are sampled from $[0,h_{max}]$, where $h_{max}=0.02$ (s). The next system state $x(t_n+h(t_n))$ is interpolated between adjacent $x$ points. The control function $u_{(t_0,t_n]}$, time spacing $h(t_n)$, and current system state $x(t_n)$ are the inputs of neural network. The next system state $x(t_n+h(t_n))$ is the output. In total, we have $N_{train}=50000$, i.e., 50000 sub-sequences of $u(t)$ with various initial values and lengths, corresponding to 50000 sets of current and next states.\\
The trained LSTM-MIONet is used to predict the response of the non-autonomous pendulum system over the time-domain $[0,10]$ (s), with random initial conditions $x(0) \in \mathcal{X}_0$, with a fixed step size of $\Delta = 0.01$ (s) and the fixed time spacing $h=0.01$ (s).\\
To test the predictive power of the proposed framework, we compute the average and standard deviation (st. dev.) of $L_2$ of 100 trajectories of the pendulum system. We sample the control functions of those trajectories from Gaussian random field $u \sim \mathcal{G}$ and initial conditions from $\mathcal{X}_0$. Then, We replicate each $u(t)$ 200 times to generate 200 sub-sequences with varying lengths, so the $L_2$ relative error is calculated at 200 state points on each $x(t)$ trajectory. In total, we have the $L_2$ vector of size 100 for all test functions. Results are reported in Table~\ref{tab:tb2}.\\
To test the predictive power of the proposed framework on out-of-distribution (dist.) datasets, we sample 100 random initial conditions sampled from $\mathcal{X}_0$, but instead of sampling the control functions from $\mathcal{G}$, we let $u(t)=\sin(t/2)$. Then, we compute the average and standard deviation of the $L_2$ of the 100 trajectories. Results are reported in Table~\ref{tab:tb2}.\\
Figure~\ref{fig:3} shows the comparison between the LSTM-enhanced MIONet prediction and the true response of the pendulum system's state variables $\theta(t)$ and $\dot{\theta}(t)$. The results demonstrate excellent agreement between the proposed method and the true values, despite the out-of-distribution nature of the control function.\\
The results of UQ using reSGLD are shown in Figure~\ref{fig:UQ-pendulum}. The number of ensembles is $M=300$. As shown in Table~\ref{tab:UQ-pendulum}, the $L_2$ relative error of the mean prediction versus the true trajectory is 3.34~\%. The PICP is 100~\%, i.e., the confidence interval covers all the true values. 

\begin{table}[ht]
    \centering
    \begin{tabular}{p{3cm}p{3cm}p{4cm}}
        \toprule
         & in dist. & out-of-dist.  \\
        \midrule
         mean $L_2$ & 2.02\% & 2.88\%\\
         st.dev. $L_2$ & 1.46\% & 1.28\%\\
         \bottomrule
    \end{tabular}
    \caption{The average and standard deviation (st.dev.) of the $L_2$ - relative error between the predicted and actual response trajectories of the pendulum system's state response $\theta(t)$ to  100 initial conditions sampled from the set $\mathcal{X}_0 \colon = \{ \theta,\dot{\theta} \colon \theta \in [-\pi,\pi], \dot{\theta} \in [-8,8] \}$, along with (i) the input signal $u \sim \mathcal{G}$ and (ii) $u=\sin(t/2)$ within the partition $\mathcal{P} \in [0,10]$ (s) of constant time spacing $h=0.01$ (s).}
    \label{tab:tb2}
\end{table}
\begin{table}[ht]
    \centering
    \begin{tabular}{p{3cm}p{2cm}}
        \toprule
         & $\theta(t)$ \\
        \midrule
         $L_2$ & 3.34\% \\
         PICP & 100.00\% \\
         \bottomrule
    \end{tabular}
    \caption{The $L_2$ relative error and PICP of the example ensemble results of pendulum system.}
    \label{tab:UQ-pendulum}
\end{table}
\subsection{A Power Engineering Application}
The evolution of residential energy systems, driven by the widespread adoption of small-scale solar photovoltaic (PV) systems, presents a new paradigm that is significantly different from traditional energy systems. This shift necessitates a blend of traditional energy consumption models and \textit{data-driven} models of solar PV generation for accurate prediction and understanding of future energy consumption and generation patterns. In this context, the Ausgrid dataset \cite{ratnam2017residential}, which provides detailed load and rooftop PV generation data for 300 de-identified residential customers in an Australian distribution network, emerges as a critical resource. The dataset reported measurements of both the load and PV generation at \textit{30-minute} intervals for 36 months, from July-1-2010 to June-30-2013. \\
In this example, we focus on the \textit{Gross Generation} (GG) data from the Ausgrid dataset, which represents the power generated by the PV systems. Unlike residential load data, which can be significantly influenced by resident habits, PV generation is predominantly affected by environmental factors such as weather conditions and geographical location. Therefore, accurate prediction of GG data can provide valuable insights for power generation planning. Note that the PV generation can be modeled as a non-autonomous system, by assuming the input function corresponds to the GG data.\\
\textit{Training Data}. We generated training data $\mathcal{D}_{train}$ using the GG data of customers 1-50 collected from June-30-2010 to July-1-2011. Each GG record corresponds to the PV power generation of a specific customer on a specific day, thus we have $366 \times 50 = 18300$ trajectories. Each trajectory contains 24 hours of data measured at \textit{30-min} interval, resulting in 48 data points. Note that the PV power generation relies on sunlight, so we only use the data points indexed from 18 to 38, corresponding to 9 am to 7 pm of a day. We use 9 am, or the $18^{th}$ data point, as the starting point of each trajectory, so the times $t$ of trajectories correspond to the uniform partition $\mathcal{P} \in [0,10]$ (hours). Then, We interpolate the data on $\mathcal{P}$ with a constant time spacing of $h = 0.05$ (hours), which gives 210 data points for each row of data. We denote the interpolated data using $x_m(t)$. Next, to produce sub-sequences of $x_m$, i.e., $x_{m}^n$, we replicate $x_m$ for 5 times and apply the mask $mask^n=[t_0<t<t_n]$ to each $x_m$, where $t_0=0$ and $t_n$ is uniformly taken from $(0,19.5)$ (hours). The local step sizes $h_n$ are sampled from $[0,0.5]$ (hours). The next system state $x(t_{n+1}) \equiv x(t_{n}+h_n)$ is interpolated between adjacent $x_m$ points. The masked solution function $x_{m}^n$, local step size $h_n$, and current system state $x(t_n)$ are taken as inputs of the training. The next system state $x(t_{n+1})$ is the output. In total, we have $N_{train}=366 \times 50 \times 5=91500$, i.e., 91500 sub-sequences of $x_m(t)$ with various initial values and lengths, corresponding to 91500 sets of current and next states.\\
The trained DeepONet is used to predict the response of the autonomous power generation system over the time-domain $[0,10]$ (hours), with a fixed step size of $h=0.05$ (hours). The local step size corresponding to the position of the next state is $h_n=0.25$ (hours).\\
To test the predictive power of the proposed framework, we compute the average and standard deviation (st. dev.) of $L_2$ of trajectories corresponding to customer 51-60 and customer 61-70, within the time from July-1-2010 to June-30-2011. We replicate each $x_m(t)$ 100 times to generate 100 sub-sequences with varying lengths, so the $L_2$ relative error is calculated at 100 state points on each $x_m(t)$ trajectory. In total, we have the $L_2$ vector of size $366 \times 10=3660$ for the two customer groups, respectively. Results are reported in Table~\ref{tab:tb5}. A comparison of the proposed DeepONet prediction with the actual trajectory of the power generation system is presented in Figure~\ref{fig:4}.\\
\begin{figure}[h!] 
    \begin{subfigure}[!bl]{0.45\textwidth}
        \begin{tikzpicture}[spy using outlines={magnification=2, width=1.5cm, height=4.5cm, connect spies}]
        \begin{axis}[width=\linewidth, xlabel={$t$ (hours)},ylabel={$x$ (kWh)}, xmin=0, xmax=10.5, ymin=0, ymax=0.6, 
        xtick={0,2,...,10}, ytick={0,0.1,...,0.6}, legend pos=north east, legend style={font=\tiny}, legend cell align={left}]
        \addplot [smooth, red] table{data/Ausgrid_infer_cust_51_true.txt}; 
        \addlegendentry{True}
        \addplot [dashed, blue] table{data/Ausgrid_infer_cust_51_pred.txt}; 
        \addlegendentry{LSTM-MIONet}
        \end{axis}
        \end{tikzpicture}
        \label{subfig:cust_51}
    \end{subfigure}
    \hspace{5pt}
    \begin{subfigure}[!bl]{0.45\textwidth}
        \begin{tikzpicture}[spy using outlines={magnification=2, width=1.5cm, height=4.5cm, connect spies}]
        \begin{axis}[width=\linewidth, xlabel={$t$ (hours)},ylabel={$x$ (kWh)}, xmin=0, xmax=10.5, ymin=0, ymax=0.5, 
        xtick={0,2,...,10}, ytick={0,0.1,...,0.5}, legend pos=north east, legend style={font=\tiny}, legend cell align={left}]
        \addplot [smooth, red] table{data/Ausgrid_infer_cust_60_true.txt}; 
        \addlegendentry{True}
        \addplot [dashed, blue] table{data/Ausgrid_infer_cust_60_pred.txt}; 
        \addlegendentry{LSTM-MIONet}
        \end{axis}
        \end{tikzpicture}
        \label{subfig:cust_60}
    \end{subfigure}\\
    \begin{subfigure}[!bl]{0.45\textwidth}
        \begin{tikzpicture}[spy using outlines={magnification=2, width=1.5cm, height=4.5cm, connect spies}]
        \begin{axis}[width=\linewidth, xlabel={$t$ (hours)},ylabel={$x$ (kWh)}, xmin=0, xmax=10.5, ymin=0, ymax=0.5, 
        xtick={0,2,...,10}, ytick={0,0.1,...,0.5}, legend pos=north east, legend style={font=\tiny}, legend cell align={left}]
        \addplot [smooth, red] table{data/Ausgrid_infer_cust_70_true.txt}; 
        \addlegendentry{True}
        \addplot [dashed, blue] table{data/Ausgrid_infer_cust_70_pred.txt}; 
        \addlegendentry{LSTM-MIONet}
        \end{axis}
        \end{tikzpicture}
        \label{subfig:cust_70}
    \end{subfigure}
    \hspace{5pt}
    \begin{subfigure}[!bl]{0.45\textwidth}
        \begin{tikzpicture}[spy using outlines={magnification=2, width=1.5cm, height=4.5cm, connect spies}]
        \begin{axis}[width=\linewidth, xlabel={$t$ (hours)},ylabel={$x$ (kWh)}, xmin=0, xmax=10.5, ymin=0, ymax=0.5, 
        xtick={0,2,...,10}, ytick={0,0.1,...,0.5}, legend pos=north east, legend style={font=\tiny}, legend cell align={left}]
        \addplot [smooth, red] table{data/Ausgrid_infer_cust_80_true.txt}; 
        \addlegendentry{True}
        \addplot [dashed, blue] table{data/Ausgrid_infer_cust_80_pred.txt}; 
        \addlegendentry{LSTM-MIONet}
        \end{axis}
        \end{tikzpicture}
        \label{subfig:cust_80}
    \end{subfigure}
\caption{Comparison of LSTM-MIONet prediction with the actual trajectory of the power generation system’s state $x(t)$ of customer 51 (\textit{upper left}), customer 60 (\textit{upper right}), customer 70 (\textit{bottom left}), and customer 80 (\textit{bottom right}), on July-1-2010 within the partition $\mathcal{P} \in [0,10]$ (hours) of constant time spacing $\Delta = 0.05$ (hours). The prediction spacing is $h=0.25$ (hours).}
\label{fig:4}
\end{figure}
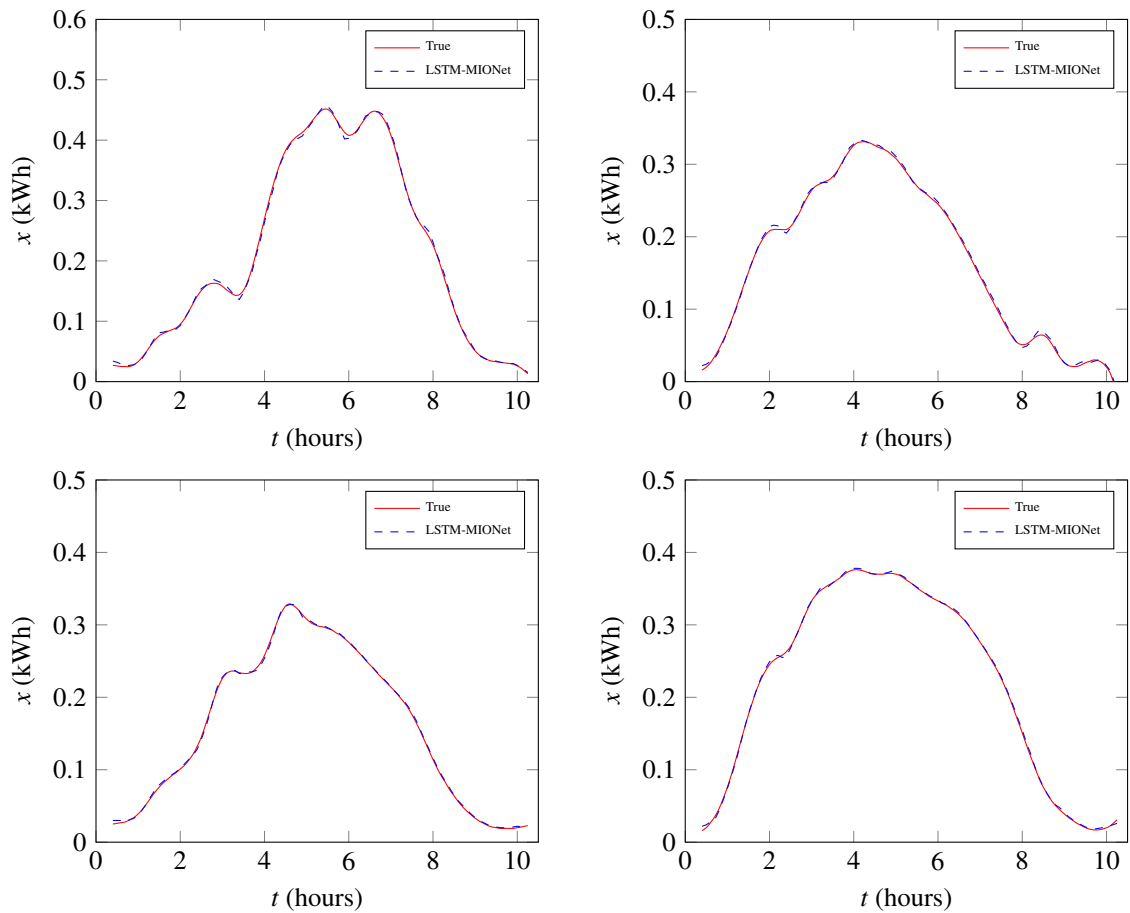
\begin{table}[ht]
    \centering
    \begin{tabular}{p{3cm}p{3cm}p{4cm}}
        \toprule
         & customer 51-60 & customer 61-70  \\
        \midrule
         mean $L_2$ & 1.23\% & 1.33\%\\
         st.dev. $L_2$ & 0.67\% & 0.73\%\\
         \bottomrule
    \end{tabular}
    \caption{The average and standard deviation (st.dev.) of the $L_2$ - relative error between the predicted and actual response trajectories of the power generation system's state response $x(t)$ of 20 customers in a year from July-1-2010 to June-30-2011. }
    \label{tab:tb5}
\end{table}

\section{Discussion} \label{sec:temporal}
\subsection{The Comparison with Existing Schemes}
In this section, we will compare the results of the numerical experiments with two existing machine learning schemes, (i) the local DeepONet \cite{lin2023learning}, and (ii) the vanilla LSTM neural network.\\
The local DeepONet \cite{lin2023learning} takes the current system state, time spacing, and the next-step control as the inputs of \textit{Non-Autonomous} system. The inputs for \textit{Autonomous} system exclude the next-step control vector, so the training dataset format becomes:
\begin{align*}
    \mathcal{D}_{train}=
    \begin{cases} 
    \{x^i(t_n),u^i{(t_n+h^i(t_n))},h^i(t_n),x^i(t_n+h^i(t_n))\}_i^{N_{train}} & 
    \small{\text{\textit{for Non-Autonomous Systems}}} \\
    \{x^i(t_n),h^i(t_n),x^i(t_n+h^i(t_n))\}_i^{N_{train}} & \small{\text{\textit{for Autonomous Systems}}}
    \end{cases}
\end{align*}
The vanilla LSTM neural network encodes the history of input functions but does not encode the time spacing or the current system state. Its training dataset format follows: 
\begin{align*}
    \mathcal{D}_{train}=
    \begin{cases} 
    \{u^i_{(t_0,t_n]},x^i(t_n+h)\}_i^{N_{train}} & 
    \small{\text{\textit{for Non-Autonomous Systems}}} \\
    \{x^i_{(t_0,t_n]},x^i(t_n+h)\}_i^{N_{train}} & 
    \small{\text{\textit{for Autonomous Systems}}}
    \end{cases}
\end{align*}

\begin{table}[ht]
    \centering
    \begin{tabular}{p{3cm}p{3cm}p{3cm}p{3cm}}
        \toprule
         & LSTM-MIONet & Local DeepONet \\
        \midrule
         mean $L_2$ & 2.02\% & 2.04\% \\
         st.dev. $L_2$ & 1.46\% & 1.50\%\\
         \bottomrule
    \end{tabular}
    \caption{The average and standard deviation (st.dev.) of the $L_2$ - relative error between the predicted and actual response trajectories of the pendulum system. }
    \label{tab:compare-schemes-Lorentz}
\end{table}
\begin{table}[ht]
    \centering
    \begin{tabular}{p{3cm}p{3cm}p{3cm}p{3cm}}
        \toprule
         & LSTM-MIONet & DeepONet\\
        \midrule
         mean $L_2$ & 1.29\% & 5.23\% \\
         st.dev. $L_2$ & 0.93\% & 0.78\%\\
         \bottomrule
    \end{tabular}
    \caption{The average and standard deviation (st.dev.) of the $L_2$ - relative error between the predicted and actual response trajectories of the Lorentz system. }
    \label{tab:compare-schemes-pendulum}
\end{table}

\subsection{The Effect of Temporal Parameters}
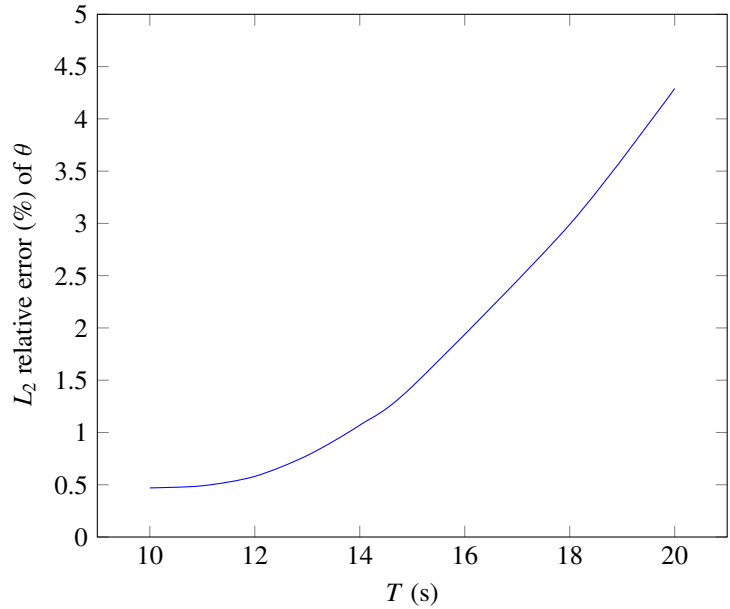
\begin{figure}[h!]
\centering
    \begin{tikzpicture}    
    \begin{axis}[width=0.6\linewidth, xlabel={$T$ (s)},ylabel={$L_2$ relative error (\%) of $\theta$}, xmin=9, xmax=21, ymin=0, ymax=5, 
    xtick={10,12,...,20}, ytick={0,0.5,...,5}, legend pos=north east, legend style={font=\small}, legend cell align={left}]
    \addplot [smooth, blue] table{data/infer_T.txt}; 
    \end{axis}
    \end{tikzpicture}
\caption{Comparison of LSTM-MIONet prediction of the non-autonomous pendulum system’s state $\theta(t)$ for (i) $u \sim \mathcal{G}$ within the partition $\mathcal{P} \in [0,T]$ (s) of constant time spacing $h = 0.01$, $T \in [10,20]$.}
\label{fig:5}
\end{figure}

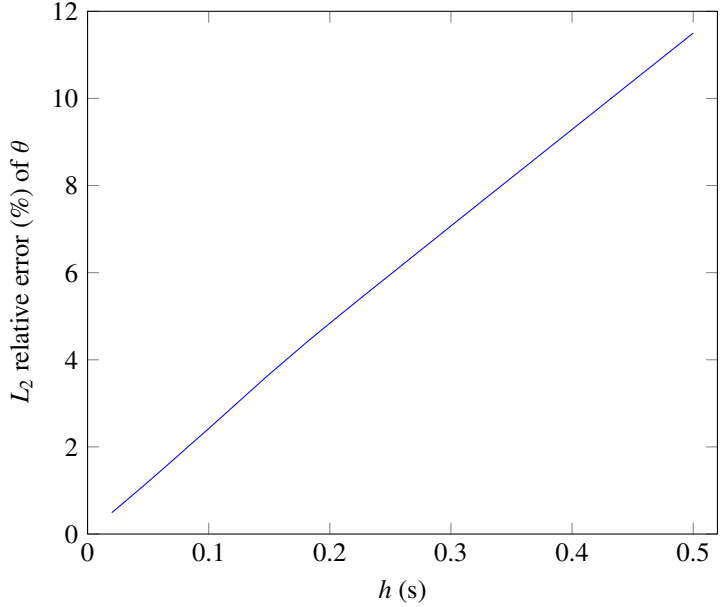
\begin{figure}[h!]
\centering
    \begin{tikzpicture}    
    \begin{axis}[width=0.6\linewidth, xlabel={$h$ (s)},ylabel={$L_2$ relative error (\%) of $\theta$}, xmin=0, xmax=0.52, ymin=0, ymax=12, 
    xtick={0,0.1,...,0.5}, ytick={0,2,...,12}, legend pos=north east, legend style={font=\small}, legend cell align={left}]
    \addplot [smooth, blue] table{data/infer-in-dist-hn.txt}; 
    \end{axis}
    \end{tikzpicture}
\caption{Comparison of LSTM-MIONet prediction of the non-autonomous pendulum system’s state $\theta(t)$ for (i) $u \sim \mathcal{G}$ within the partition $\mathcal{P} \in [0,10]$ (s) of constant time spacing $h \in [0.02,0.50]$.}
\label{fig:6}
\end{figure}

In the numerical experiments, we introduced the test results of the proposed framework when the time partition and step size are the same as those of the training dataset. To further explore the effect of temporal parameters on the prediction performance, we use the pendulum system trained on $t \in [0,10]$ (s) with maximum step size $h=0.02$ (s) as the benchmark. Then, we study (i) the extrapolation effect by testing the model in a time partition that is different from that of the training dataset, and (ii) the step size effect by testing the model with a different step size than that of the training dataset.\\
\textit{The Extrapolation Effect}. The aim of this experiment is to learn the dynamics of the system over a time range $t \in [0, T]$, where $T \in [10,20]$, with the fixed step size $h=0.01$. The models are trained on the time partition $[0,10]$ and tested on $[0,20]$. Results are presented in Figure~\ref{fig:5}. The $L_2$ relative error increases from 0.47\% at $T=10$ (s) to 4.29\% at $T=20$ (s). It remains small ($\leq0.58\%$) when $T\leq12$ (s), indicating the trained network can be extrapolated to 120\% of the maximum time of its training set.\\
\textit{The Step Size Effect}. In the example of the pendulum system, the step size $h$ of the test dataset was set to be small, i.e., $h=0.01$ (s). The effect of applying a larger step size to the test dataset is explored here. Results are presented in \ref{fig:6}. It is worth noting that the proposed operator learning scheme solely relies on historical data, so a higher prediction error is expected when the step size increases. Results in Figure~\ref{fig:6} indicate a linear relation of the $L_2$ relative error and the step size $h$ when $h \in [0.02,0.50]$.

\section{Conclusion} \label{sec:conclusion}
In this paper, we have explored the approximation of the dynamic responses of complex systems under varying input lengths and initial conditions. Our approach has been to integrate Bayesian methods into a novel variant of MIONet, which we term B-LSTM-MIONet. This framework uniquely combines the temporal sensitivity of LSTM branches, capable of processing input histories of varying lengths and understanding temporal causality, with a secondary branch that incorporates the current system state as reference information. Additionally, the trunk network's ability to encode step size plays a crucial role in learning the local dynamics of system trajectories.\\
A key aspect of our study has been the application of the Bayesian M--ensemble algorithm, an innovative approach to quantifying model performance uncertainty. To demonstrate the effectiveness of our proposed B-LSTM-MIONet framework, we applied it to three different numerical examples: the Lorentz system, the pendulum system, and the PV power generation system. In each instance, the framework proved highly capable of approximating dynamic responses in complex systems with varying inputs and initial conditions. Notably, our uncertainty quantification (UQ) results revealed that the confidence intervals for predictions on noisy data successfully encompassed 100~\% of the true values, underscoring the model's robustness.\\
Furthermore, our research delved into the influence of temporal parameters on model performance, focusing on extrapolation capabilities over longer time horizons and the impact of larger step sizes in the test dataset. We found that the proposed scheme could effectively extrapolate the trained network to cover 120~\% of the sequence length encountered during training. However, step size emerged as a critical factor affecting prediction accuracy. This is attributed to our framework's reliance on historical data; states that are significantly distant from the current known state exhibit a weaker connection with the trajectory's history. This finding highlights the importance of considering step size and historical data relevance when applying our B-LSTM-MIONet in practical scenarios.\\
Our future work aims to explore the synergies between Recurrent Neural Networks (RNN) and Fourier Neural Operators (FNO). The FNO's ability to learn system dynamics from a frequency domain perspective could potentially enhance prediction accuracy over longer time horizons, offering a promising avenue for further research and development in this field.

\section*{Acknowledgments}
Guang Lin, Na Lu, Christian Moya, and Amirhossein Mollaali gratefully acknowledge the support of the National Science Foundation (DMS-2053746, DMS-2134209, ECCS-2328241, and OAC-2311848), and U.S. Department of Energy (DOE) Office of Science Advanced Scientific Computing Research program DE-SC0021142, DE-SC0023161, and the Uncertainty Quantification for Multifidelity Operator Learning (MOLUcQ) project (Project No. 81739), and DOE – Fusion Energy Science, under grant number: DE-SC0024583.
\bibliographystyle{elsarticle-num} 
\bibliography{refs.bib}
\end{document}